\documentclass{article}
\usepackage[preprint]{neurips_2026}
\usepackage[utf8]{inputenc}
\usepackage[T1]{fontenc}
\usepackage{lmodern}
\usepackage{amsmath,amssymb,booktabs,array,graphicx}
\usepackage{xcolor}
\usepackage{placeins}
\usepackage{tikz}
\usetikzlibrary{positioning,arrows.meta}
\usepackage{hyperref}
\hypersetup{colorlinks=true,linkcolor=blue,citecolor=blue,urlcolor=blue,pdftitle={When to Trust Simulated Patient Responses},pdfauthor={Yifan Jin and Tao Liu}}
\newcommand{\method}{SG-WM}
\title{When to Trust Simulated Patient Responses:\\
Support-Gated World Models for Vasopressor Decision Support}
\author{Yifan Jin$^{1}$, Tao Liu$^{1}$\\
$^{1}$Sun Yat-sen University, Guangzhou, Guangdong, China}
\date{}
\makeatletter
\renewcommand{\@notice}{}
\makeatother
\begin{document}
\maketitle
\begin{abstract}
Clinical world models expose language-model agents to predicted patient
responses, but simulations can be unreliable for poorly supported treatments
or unfamiliar hospitals. We introduce Support-Gated World-Model decision
support (\method), a lightweight extension of a propose--simulate--refine
agent. A recurrent dynamics ensemble predicts short-horizon physiological
responses to documented norepinephrine dose categories, while a behavior
model estimates observational action support. The interface withholds
predictions with insufficient support or excessive ensemble disagreement
and abstains when no candidate qualifies. Evaluation on eICU-CRD includes
70 hospitals and 30,700 decision transitions, with 14 hospitals held out for
testing. At 62.0\% factual-transition coverage, joint gating yields a MAP
mean absolute error of 7.3 mmHg, compared with 9.2 mmHg for random selection
at the same coverage. The gated agent achieves 60.0\% agreement with recorded
actions at 81.0\% decision coverage, compared with 55.0\% agreement and
99.5\% coverage for an ungated ensemble agent. Component ablations show
lower factual risk with joint gating than with either gate alone.
These results characterize a reliability--coverage trade-off in selective
clinical simulation; they do not identify individualized treatment effects
or establish improved patient outcomes.
\end{abstract}

\section{Introduction}
World-model-augmented clinical agents connect a decision model to learned patient
dynamics. SepsisAgent combines treatment proposals, simulated responses, and
refinement, with additional agent training \citep{wu2026sepsisagent}.
medDreamer uses imagined latent trajectories for clinical policy learning
\citep{xu2025meddreamer}. These studies motivate asking how a decision agent
should use a simulator whose reliability varies across patients and actions.

An observational simulator has uneven data support. A candidate dose may be
uncommon for a particular history, and a new hospital may differ in case mix,
treatment documentation, and monitoring. An agent that receives all predictions
in the same format can mistake model confidence for clinical evidence.
Predictive accuracy under the recorded action also does not establish the
correctness of an unobserved treatment alternative. These problems motivate
a narrower question than autonomous treatment optimization: \emph{can a small
interface change reduce reliance on unreliable simulated responses, while
making its coverage cost measurable?}

We propose \method, which retains a recurrent clinical dynamics model and a
propose--simulate--refine agent, but adds a candidate-level support and
disagreement gate. Only eligible predictions are released to the agent.
Rejected candidates receive explicit reason codes; an empty eligible set
produces abstention. We evaluate the framework on eICU-CRD, whose multi-hospital structure
enables testing on unseen sites \citep{pollard2018eicu}. The evaluation
separates factual predictive reliability from the behavior of the
simulation-augmented agent.

The study makes three contributions: (1) a precise interface separating
supported predictions, withheld predictions, and abstention; (2) a
hospital-held-out evaluation that measures predictive error and coverage
together; and (3) a matched ensemble baseline and component ablations that
isolate the effects of support restriction and model disagreement.
On held-out hospitals, joint gating reduces factual prediction error
relative to matched random selection, while making the associated reduction
in decision coverage explicit.

\section{Related work and positioning}

\paragraph{Clinical world models.}
Generative models of longitudinal health records provide important
foundations for modeling patient trajectories. Foresight forecasts
future medical events from electronic health records, while Delphi-2M
models disease trajectories at population scale
\citep{kraljevic2024foresight,shmatko2025delphi}.
In medical imaging, MedDream learns a shared latent radiographic state
that supports diagnostic reasoning and report-conditioned image
generation, extending world-model formulations to the interpretation
and synthesis of clinical imaging evidence
\citep{xi2026radiographic}.
Intervention-oriented approaches further incorporate treatment actions
into patient-state prediction. Medical World Model explores
treatment-conditioned tumor simulation, and EHRWorld addresses
long-horizon clinical trajectory simulation under sequential
interventions \citep{yang2025mewm,mu2026ehrworld}.
Meanwhile, medDreamer and SepsisAgent integrate learned patient dynamics
into policy learning and language-agent interaction, respectively
\citep{xu2025meddreamer,wu2026sepsisagent}. Complementary medical vision-language research examines how clinical
language can be grounded in visual representations. Report-guided
segmentation translates radiology findings into spatially localized
image evidence \citep{xi2026grounding}, while pathology
vision-language pre-training combines cross-modal alignment with
self-supervised learning to strengthen visual representations
\citep{yang2025pathology}.

\paragraph{Uncertainty and conservative offline learning.}
Deep ensembles offer a practical approach to predictive uncertainty
estimation \citep{lakshminarayanan2017ensembles}, while evaluations under
dataset shift demonstrate the importance of assessing uncertainty
beyond the training distribution \citep{ovadia2019uncertainty}.
In model-based offline reinforcement learning, MOPO and MOReL address
model exploitation through uncertainty penalties and pessimistic
environment models \citep{yu2020mopo,kidambi2020morel}.
Related approaches address distributional shift through conservative
value estimation, as in Conservative Q-Learning, or restrict deviations
from a baseline policy in poorly supported regions, as in SPIBB
\citep{kumar2020cql,laroche2019spibb}.
Selective prediction provides a complementary perspective:
SelectiveNet explicitly studies the trade-off between predictive risk
and coverage through a reject option
\citep{geifman2019selectivenet}.
Our method brings these perspectives to an LLM-facing clinical
simulation interface, combining candidate-level support filtering,
disagreement-based rejection, and explicit abstention.

\paragraph{Clinical interpretation.}
Evaluating treatment recommendations from observational healthcare
data requires separating predictive performance from policy quality
and clinical benefit \citep{gottesman2019guidelines}.
Counterfactual off-policy evaluation using structural causal models
provides one framework for examining how alternative policies might
change individual trajectories under explicit modeling assumptions
\citep{oberst2019counterfactual}.
In our setting, clinician agreement measures correspondence with
recorded actions rather than treatment optimality, and historical
action frequency represents observational support rather than a
clinical safety guarantee. We therefore evaluate factual prediction
error, decision coverage, and unsupported selections separately.
These retrospective measures characterize the behavior of the
simulation interface; establishing patient benefit requires
additional causal and clinical validation.
\section{Study setting and data protocol}
\subsection{Cohort}
We study adult ICU stays with documented norepinephrine exposure in
eICU-CRD v2.0 \citep{eicu2019resource}. The analysis includes 8,650 patients,
9,010 eligible ICU stays, and 30,700 decision transitions across 70 hospitals.
The training, validation, and test partitions contain 42, 14, and 14
hospitals, respectively, and have no hospitals in common. Table~\ref{tab:cohort} summarizes cohort
characteristics and dose-record exclusions. The cohort is defined by
documented norepinephrine exposure rather than a Sepsis-3 diagnosis.

Interpretable dose records and pre-decision physiological history define
the modeling inputs. The infusion table contains heterogeneous charted
rates; missing concentration can prevent conversion from volumetric to
drug dose \citep{eicuinfusiondocs}. Unresolvable dose records are excluded,
and missing infusion records are distinguished from explicit zero rates.
The test partition includes 70 documented zero-dose transitions. Missing
MAP targets increase from 9.4\% in training to 11.0\% in testing.

\subsection{Time alignment and treatment representation}
The decision grid uses four-hour intervals within the first 72 hours
after the first norepinephrine administration record satisfying the
inclusion criteria. For each boundary $t$, history
$H_t$ uses only entries available before $t$. The index treatment $a_t$ is the
first valid documented norepinephrine rate in $(t,t+30\text{ min}]$.
The primary response $Y_{t+1}$ comprises mean arterial pressure (MAP) and
heart rate summarized over $(t+3\text{ h},t+4\text{ h}]$, with prespecified
measurement requirements. This is an index-dose-associated prediction task:
the dose is \emph{not} assumed constant throughout the four-hour interval.
Subsequent treatments and co-interventions remain potential confounders.

Rates are harmonized to $\mu$g/kg/min only when dose units and any
necessary patient weight or drug concentration are verified. The action
set consists of five discrete categories: an explicit documented zero-dose
category and four positive-dose intervals defined by the training-set
positive-dose quartiles. The cutpoints are $Q_{25}=0.05$,
$Q_{50}=0.10$, and $Q_{75}=0.20$ $\mu$g/kg/min. Unsupported or
unresolvable records are excluded. These training-derived cutpoints
remain fixed during validation and testing.

History includes prior vital-sign summaries, last available laboratory
values, measurement ages, missingness indicators, verified previous
medication and treatment records, and static covariates. Discharge information, future measurements, completed
first-day severity summaries, and future treatment summaries are prohibited
as decision inputs. Lab availability is handled conservatively using
documented result/revision offsets
\citep{eiculabdocs}.

\subsection{Hospital-held-out evaluation}
The hospital split comprises 42 training sites, 14 validation sites, and
14 test sites, corresponding to a 60/20/20 partition. The test set contains
1,800 unique patients and 6,300 decision transitions. Feature scaling,
dose-bin cutpoints, model fitting, and interface thresholds are development
operations; test hospitals serve only for evaluation. This split examines
transport to hospitals excluded from training rather than within-hospital
patient interpolation.

\section{Support-gated simulation interface}
\subsection{Shared recurrent dynamics ensemble}
For a patient history $H_t$, a two-layer GRU produces a hidden representation
$h_t$. A candidate action is embedded and concatenated to $h_t$ before
prediction. The recurrent backbone provides a lightweight history-aware representation
for the action-conditioned prediction head. Each of $M=3$ independently initialized models
is trained on a patient-bootstrap sample. Every GRU has 128 hidden units
and dropout of 0.1:
\begin{equation}
 p_{\theta_m}(Y_{t+1}\mid H_t,a_t)
 =\prod_{j=1}^{d}\mathcal{N}\bigl(Y_{t+1,j};
 \mu_{mj}(H_t,a_t),\sigma^2_{mj}(H_t,a_t)\bigr).
\end{equation}
Training minimizes Gaussian negative log likelihood on observed targets only.
Missing targets are masked rather than imputed as outcomes. Each member is
optimized using AdamW with a learning rate of $10^{-3}$, batch size 128,
and at most 100 epochs; early stopping uses a patience of 10 epochs.
The prespecified random seeds for stability assessment are 13, 42, and
2026. Predictors use training-derived scaling, and evaluation metrics
are also reported in physical units where applicable.

The ensemble mean and disagreement score are
\begin{align}
 \bar{\mu}_j(H,a)&=\frac{1}{M}\sum_{m=1}^{M}\mu_{mj}(H,a),\\
 u(H,a)&=\frac{1}{d}\sum_{j=1}^{d}
 \frac{\frac{1}{M}\sum_{m=1}^{M}
 [\mu_{mj}(H,a)-\bar{\mu}_j(H,a)]^2}{s_j^2+\epsilon},
\end{align}
where $s_j$ is the training-set target standard deviation.
Disagreement is a heuristic signal of model instability; it is not a
complete estimate of epistemic uncertainty or counterfactual error.
Total predictive variance additionally includes the average within-model
variance. Any interval rescaling is fitted only on calibration sites.
No distribution-free coverage guarantee is claimed across hospitals.

\subsection{Observational action support}
A separately trained GRU classifier estimates
$b_\phi(a\mid H)$ from training histories and recorded actions. It is
optimized with cross-entropy; its output probabilities are evaluated and
used for support-threshold calibration on the validation hospitals. Its probabilities operationalize local observational support:
\begin{equation}
 q(H,a)=b_\phi(a\mid H).
\end{equation}
Low $q$ can indicate extrapolation risk, but high $q$ does not establish
safety, clinical appropriateness, or exchangeability.
The classifier shares the feature schema, not the fitted parameters, with
the dynamics ensemble.

\subsection{Candidate gate and abstention}
The agent proposes up to three distinct action-bin identifiers from the
predefined action set. For each candidate, the interface computes
\begin{equation}
 G(H,a)=\mathbb{I}\{q(H,a)\geq\eta\}
         \mathbb{I}\{u(H,a)\leq\tau\}.
\end{equation}
The support threshold $\eta$ and disagreement threshold $\tau$ define
the acceptance region. Starting from $\eta_0=0.05$ and $\tau_0=0.03$,
validation-set quantile calibration and grid search select the final
thresholds $\eta^{*}=0.06$ and $\tau^{*}=0.025$. Calibration uses
recorded actions from the validation partition; the selected thresholds
remain fixed throughout test-hospital evaluation. These are statistical
interface thresholds rather than clinical risk limits.

When $G=1$, the tool returns the mean response, predictive spread,
disagreement score, and support score. When $G=0$, it withholds the
numerical prediction and returns \texttt{LOW\_SUPPORT},
\texttt{HIGH\_DISAGREEMENT}, or both.
The language-model agent may select only a candidate with $G=1$; a
deterministic validator enforces this restriction. It abstains if no
candidate qualifies. Uncertainty feedback and enforcement are distinct
mechanisms.

\subsection{Proposal and refinement configuration}
We use Qwen2.5-7B-Instruct for the proposal and refinement stages.
Decoding is deterministic, with temperature 0, top-$p$ 1.0, and a
maximum output length of 256 tokens per generation. At most three
distinct candidate dose categories are proposed at each decision point.
Responses follow a structured JSON schema containing action-category
identifiers. A single repair attempt is permitted after a malformed
response; another parsing failure results in invalid-output abstention.
The deterministic validator checks the final category against the
eligible set and rejects any selection that violates the support or
disagreement constraints. The configuration and threshold-selection
protocol are summarized in Appendix~\ref{app:repro}.

\begin{figure}[t]
\centering
\begin{tikzpicture}[
 node distance=5mm,
 box/.style={draw,rounded corners,align=center,text width=10.4cm,
 inner sep=5pt,font=\small},
 arr/.style={-{Latex[length=2mm]},thick}]
 \node[box] (history) {Available patient history $H_t$};
 \node[box,below=of history] (proposal) {Shared LLM: propose up to three dose categories};
 \node[box,below=of proposal] (scores) {Shared dynamics ensemble + behavior model:
 response, disagreement, support};
 \node[box,below=of scores] (gate) {Candidate gate:
 release supported predictions; withhold rejected predictions};
 \node[box,below=of gate] (final) {Shared LLM: refine choice within eligible set\\
 Validator: accept eligible choice or record abstention};
 \draw[arr] (history)--(proposal);
 \draw[arr] (proposal)--(scores);
 \draw[arr] (scores)--(gate);
 \draw[arr] (gate)--(final);
\end{tikzpicture}
\caption{Support-gated simulation interface. Candidate predictions are released only when they satisfy both observational-support and model-disagreement constraints.}
\label{fig:architecture}
\end{figure}
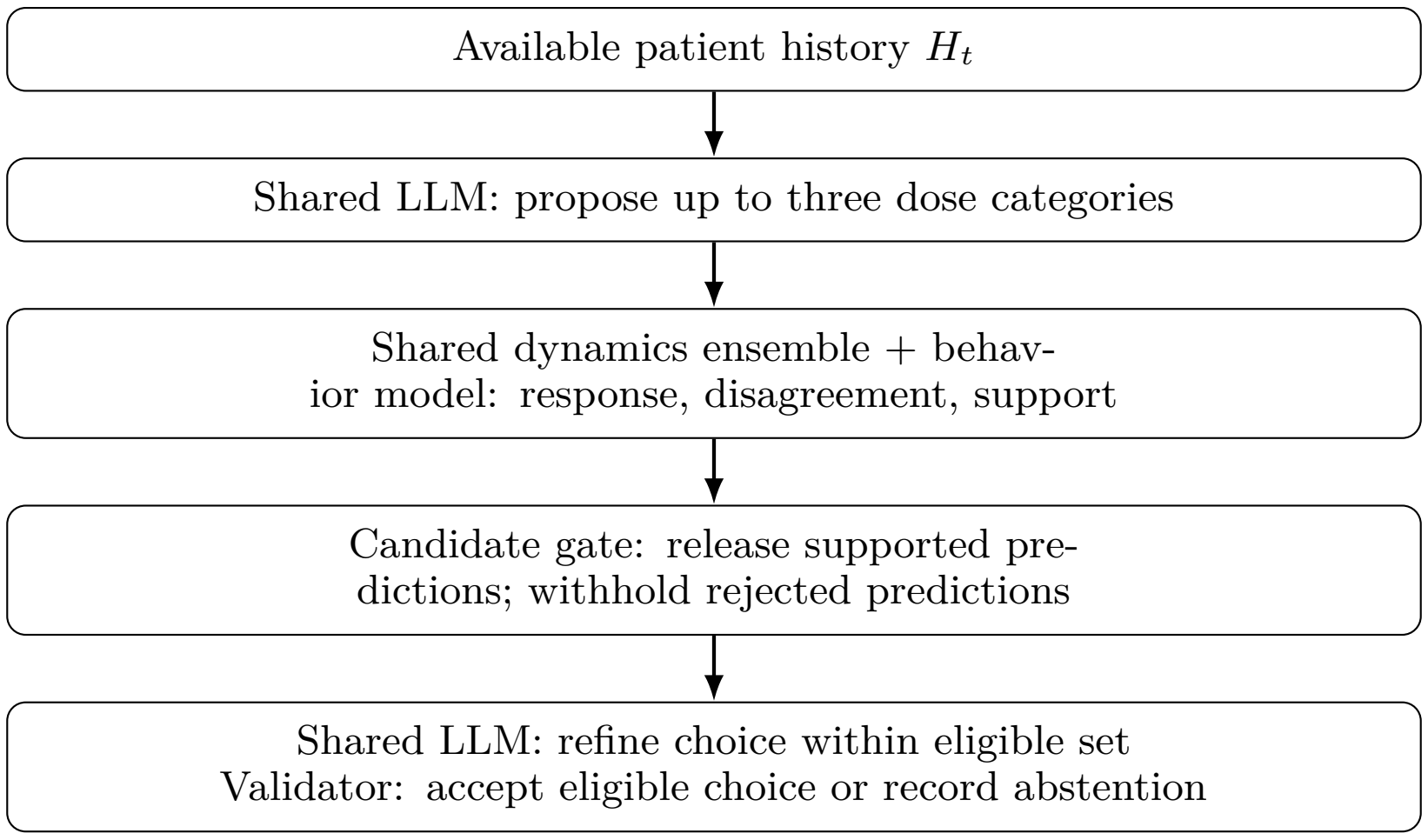

\subsection{Matched baseline and scope of adaptation}
The principal comparator is an \emph{ensemble world-model agent without
gating}. It uses the same patient histories, candidate proposals, dynamics
ensemble, LLM checkpoint, tool budget, and response summaries. It receives
predictions for all proposed candidates and selects one without hard
support/disagreement filtering. Matching ensemble size prevents attributing
an ensembling benefit to the gate.

The comparator is a SepsisAgent-inspired inference baseline using the
same proposal--simulation--refinement structure. The comparison isolates
the inference interface rather than the source paper's complete
three-stage agent-training pipeline.

\section{Experimental evaluation}
\subsection{Comparators and evaluation scope}
We compare a behavior-cloning classifier, a direct LLM without simulation,
an ungated ensemble world-model agent, and \method. The ungated and gated
agents share the dynamics ensemble and candidate-generation structure.
The principal comparison therefore concerns access to candidate-level
simulation feedback. Calls per case summarize average inference usage.

The evaluation separates two units of analysis. Factual prediction assesses
a simulator under the action recorded in the data. Agent evaluation
assesses whether a candidate set produces a non-abstaining output and
which action category is selected. A factual transition may fail the
gate while the agent still has another eligible candidate. Consequently,
factual coverage and agent decision coverage have different denominators.

\subsection{Factual prediction metrics}
For a recorded action, normalized absolute prediction error is
\begin{equation}
 e_i=\frac{1}{|O_i|}\sum_{j\in O_i}
 \frac{|Y_{ij}-\bar{\mu}_j(H_i,a_i)|}{s_j+\epsilon},
\end{equation}
where $O_i$ indexes observed target dimensions and $s_j$ is the
training-set standard deviation of target $j$.
For accepted factual transitions $A=\{i:G(H_i,a_i)=1\}$,
\begin{equation}
 \operatorname{Coverage}=\frac{|A|}{N},\qquad
 \operatorname{Risk}=\frac{1}{|A|}\sum_{i\in A}e_i.
\end{equation}
Physical-unit MAP and heart-rate MAE provide interpretable measures of
predictive error. Prediction-interval (PI) coverage measures the
proportion of observed targets contained in the nominal 90\% interval.
The joint gate is compared with support-only selection,
disagreement-only selection, and random selection at the same factual
coverage. Metrics are evaluated only where the corresponding target is
observed. Factual prediction does not validate responses to unrecorded
treatment alternatives.

\subsection{Agent metrics and component ablation}
Agent coverage is the proportion of evaluated cases with a non-abstaining
output. Agreement is the proportion of non-abstaining decisions for which the
selected action category matches the recorded category. Unsupported-
selection rate is the proportion of selected actions that fail the
behavior-support criterion $q(H,a)\geq\eta$. We additionally report
average calls per case. Agreement is descriptive and does not establish
that recorded clinician actions are optimal. A zero unsupported-selection
rate is partly enforced by the interface and is not independent evidence
of clinical safety.

The component ablation compares no gate, the support gate alone, the
disagreement gate alone, and the joint gate, using a common ensemble.
We interpret factual risk together with agent coverage. Because the
reported results are aggregate point estimates without per-seed dispersion
or hospital-level uncertainty estimates, comparisons are descriptive;
no statistical significance or confidence interval is inferred from
the tabulated values.

\section{Results}
\subsection{Cohort composition}
The hospital-held-out test partition contains 1,880 ICU stays and 6,300
decision transitions. Its 11.0\% MAP-target missingness exceeds the 9.4\%
observed in training. Across the three partitions, 12,500 unresolvable
dose records are excluded. These counts describe record exclusions
rather than additional excluded patients.

\begin{table}[ht]
\centering\small
\caption{Cohort characteristics by hospital partition. Dose exclusions are record counts.}
\label{tab:cohort}
\begin{tabular}{lrrr}
\toprule
Characteristic & Training & Validation & Test\\
\midrule
Hospitals & 42 & 14 & 14\\
Unique patients & 5,200 & 1,650 & 1,800\\
Eligible ICU stays & 5,400 & 1,730 & 1,880\\
Decision transitions & 18,500 & 5,900 & 6,300\\
Documented zero-dose transitions & 210 & 65 & 70\\
Missing MAP targets (\%) & 9.4 & 10.1 & 11.0\\
Excluded unresolvable dose records & 7,300 & 2,400 & 2,800\\
\bottomrule
\end{tabular}
\end{table}

\subsection{Selective prediction on unseen hospitals}
Joint gating retains 62.0\% of factual transitions, with MAP MAE of
7.3 mmHg and heart-rate MAE of 9.6 bpm (Table~\ref{tab:prediction}).
Random selection at the same coverage yields errors of 9.2 mmHg and
11.6 bpm, respectively. The corresponding relative error reductions are
20.7\% for MAP and 17.2\% for heart rate. The matched-coverage comparison
suggests that the retained transitions are more predictable than a
random subset of equal size.

The support-only gate retains 77.0\% of factual transitions with MAP MAE
of 8.6 mmHg, whereas the disagreement-only gate retains 79.0\% with MAP
MAE of 7.9 mmHg. Joint gating has the lowest error but also the lowest
coverage among these rules, so these unmatched-coverage comparisons
should not be interpreted as isolating a predictive advantage independent
of selection. Empirical coverage of nominal 90\% prediction intervals is
81.5\% without gating, 84.0\% for support-only gating, 85.7\% for
disagreement-only gating, and 87.8\% for joint gating. Coverage remains
below the nominal 90\% level for every selection rule.

\begin{table}[ht]
\centering\small
\caption{Factual prediction on unseen hospitals. Selection coverage is
reported in percent; MAP MAE is measured in mmHg and HR MAE in bpm.
Random selection is matched to the joint gate's 62.0\% coverage.}
\label{tab:prediction}
\begin{tabular}{lrrr}
\toprule
Selection rule & Coverage (\%) & MAP MAE & HR MAE\\
\midrule
All factual transitions & 100.0 & 9.2 & 11.6\\
Support only & 77.0 & 8.6 & 10.9\\
Disagreement only & 79.0 & 7.9 & 10.3\\
Joint gate & 62.0 & 7.3 & 9.6\\
Random selection (matched) & 62.0 & 9.2 & 11.6\\
\bottomrule
\end{tabular}
\end{table}

\subsection{Agent behavior and decision coverage}
The ungated ensemble agent produces a decision in 99.5\% of cases,
with 55.0\% agreement and a 12.0\% unsupported-selection rate.
\method\ achieves 60.0\% agreement at 81.0\% coverage, with no
unsupported selections (Table~\ref{tab:agent}). Relative to the ungated
agent, these values represent a 5.0 percentage-point increase in
agreement, an 18.5 percentage-point reduction in coverage, and a
12.0 percentage-point reduction in unsupported selections.
The lower coverage is an essential part of the result: 60.0\% agreement
is conditional on the 81.0\% of decision cases accepted by the gate,
not the entire test cohort. The comparison is therefore not a paired
assessment of action selection on the same patients, and it does not
establish an improvement in clinical outcomes.

Average calls per case are 4.6 for \method\ and 4.8 for the ungated
ensemble agent, a descriptive reduction of 4.2\%. These aggregate call
counts do not establish lower monetary cost or latency. The direct
language-model baseline uses fewer calls but has lower agreement and a
higher unsupported-selection rate. Behavior cloning has no unsupported
selections in the reported evaluation, showing that this diagnostic
alone does not distinguish the proposed agent.

\begin{table}[ht]
\centering\small
\caption{Agent evaluation. Coverage, agreement, and unsupported-selection
rates are percentages; agreement is evaluated on non-abstaining decisions.}
\label{tab:agent}
\resizebox{\linewidth}{!}{
\begin{tabular}{lrrrr}
\toprule
Method & Coverage (\%) & Agreement (\%) & Unsupported (\%) & Calls/case\\
\midrule
Behavior cloning & 100.0 & 53.0 & 0.0 & 0.0\\
Direct LLM & 98.0 & 48.0 & 14.0 & 1.0\\
Ungated ensemble agent & 99.5 & 55.0 & 12.0 & 4.8\\
\method & 81.0 & 60.0 & 0.0 & 4.6\\
\bottomrule
\end{tabular}}
\end{table}

\subsection{Component ablation}
As shown in Table~\ref{tab:ablation}, factual risk is 0.53 without
gating, 0.50 with support-only gating, 0.46 with disagreement-only
gating, and 0.43 with the joint gate. Joint gating reduces reported
risk by 0.10, or 18.9\%, relative to the ungated interface.
Its agent coverage is 81.0\%, compared with 93.0\% for support-only
and 90.0\% for disagreement-only gating. The two criteria therefore
produce a stronger selection effect when combined. Because the
retained cases differ, the ablation supports a risk--coverage trade-off
rather than a coverage-independent claim of superiority.

\begin{table}[ht]
\centering\small
\caption{Component ablation. Factual risk is normalized prediction error;
agent coverage is measured on candidate sets. The two metrics use
different evaluation units.}
\label{tab:ablation}
\resizebox{\linewidth}{!}{
\begin{tabular}{lccrr}
\toprule
Interface & Support gate & Disagreement gate & Factual risk $\downarrow$ & Agent coverage (\%)\\
\midrule
Ungated & No & No & 0.53 & 99.5\\
Support only & Yes & No & 0.50 & 93.0\\
Disagreement only & No & Yes & 0.46 & 90.0\\
Joint gate & Yes & Yes & 0.43 & 81.0\\
\bottomrule
\end{tabular}}
\end{table}
\FloatBarrier

\section{Discussion and limitations}
The results support selective use of clinical simulation: at the same
62.0\% factual coverage, joint gating identifies transitions with lower
MAP and heart-rate error than random selection. The interval-coverage
increase is consistent with a more predictable retained subset,
although undercoverage remains. Combining action support and ensemble
disagreement also yields lower factual risk than either criterion alone,
at the cost of more abstention.

Agent decision coverage is higher than factual gate coverage because an
agent can choose among several proposed candidates, whereas factual
evaluation applies the gate to the single recorded action.
The 81.0\% agent coverage therefore does not conflict with 62.0\%
factual coverage. At the same time, higher action agreement on accepted
cases cannot establish improved treatment selection for the full
population. Unsupported-selection rates measure a property targeted
by the gate and must be interpreted with that construction in mind.

The analysis has several limitations. Aggregate estimates do not quantify
sampling variability, seed-to-seed variation, or hospital-specific
performance. An ensemble may agree while being systematically wrong,
and a behavior model may reproduce biased
historical practice. Selective rejection can disproportionately affect
patients with sparse monitoring or unusual treatment histories; subgroup
error and coverage are needed to characterize this risk.

The index-dose formulation simplifies a continuously changing treatment
process. Delayed documentation, concurrent drugs, fluids, severity, and
clinician judgment can confound associations between dose and future
physiology. Dose binning discards detail. The small number of documented
zero-dose transitions further limits evidence about withholding or
discontinuing norepinephrine. Excluding unresolvable records narrows
the represented population, and hospital-held-out evaluation does not
address all temporal or geographic shifts. These findings concern
retrospective factual prediction and interface behavior rather than
causal treatment effects, mortality benefit, or prospective safety.

\section{Conclusion}
\method\ combines observational action support and ensemble disagreement
to regulate a clinical agent's access to simulated patient responses.
On held-out eICU hospitals, the joint gate yields a MAP MAE of 7.3 mmHg
at 62.0\% factual coverage, compared with 9.2 mmHg for matched random
selection. The gated agent achieves 60.0\% recorded-action agreement
at 81.0\% decision coverage, while enforcing its support constraint.
The main empirical finding is a measurable reliability--coverage
trade-off, motivating validation of selective simulation interfaces
before their use in clinical decision support.

\bibliographystyle{plainnat}
\bibliography{references}
\clearpage
\appendix
\section{Implementation and calibration protocol}
\label{app:repro}
Table~\ref{tab:configuration} summarizes the dynamics ensemble,
agent, and validation-only gate configuration.

\begin{table}[ht]
\centering\small
\caption{Model, decoding, action discretization, and gate-configuration parameters.}
\label{tab:configuration}
\begin{tabular}{p{0.40\linewidth}p{0.52\linewidth}}
\toprule
Component & Configuration\\
\midrule
Dynamics ensemble & 3 independently initialized GRUs; 2 layers per GRU; 128 hidden units; dropout 0.1\\
Dynamics optimizer & AdamW; learning rate $10^{-3}$; batch size 128; at most 100 epochs; early-stopping patience 10\\
Stability-assessment seeds & 13, 42, 2026\\
Behavior support estimator & Separate GRU classifier with categorical cross-entropy loss\\
Action representation & Documented zero dose plus 4 positive-dose bins; training cutpoints $Q_{25}=0.05$, $Q_{50}=0.10$, $Q_{75}=0.20$ $\mu$g/kg/min\\
Proposal and refinement model & Qwen2.5-7B-Instruct\\
Decoding & Temperature 0; top-$p$ 1.0; 256 maximum output tokens\\
Candidate generation & At most 3 distinct candidate actions per decision\\
Output validation & Structured JSON; at most 1 format-repair attempt; deterministic eligible-action check\\
Initial gate thresholds & Support $\eta_0=0.05$; disagreement $\tau_0=0.03$\\
Final gate thresholds & Support $\eta^{*}=0.06$; disagreement $\tau^{*}=0.025$\\
Threshold tuning & Quantile calibration and grid search on validation hospitals only; final thresholds fixed for test evaluation\\
\bottomrule
\end{tabular}
\end{table}

\paragraph{Dataset partition and preprocessing.}
The development protocol partitions 70 hospitals into 42 training,
14 validation, and 14 test sites. The respective patient counts are
5,200, 1,650, and 1,800; eligible ICU-stay counts are 5,400,
1,730, and 1,880; and decision-transition counts are 18,500,
5,900, and 6,300. The partitions contain 210, 65, and 70 documented
zero-dose transitions, respectively. MAP-target missingness is
9.4\%, 10.1\%, and 11.0\%. Dose-record exclusions caused by
unresolvable concentration, weight, or units number 7,300, 2,400,
and 2,800 records, for a total of 12,500. These exclusions are
record-level counts, not patient exclusions. Action bin cutpoints and
feature scaling are derived exclusively from training data.

\paragraph{Validation and test isolation.}
Support scores are estimated from past observations and the candidate
action; disagreement is computed from the three dynamic-model means.
Starting from $\eta_0$ and $\tau_0$, quantile-based calibration and
grid search on validation sites yield $\eta^{*}=0.06$ and
$\tau^{*}=0.025$. The resulting pair is locked for all test-hospital
analyses, including
the support-only, disagreement-only, and joint-gate assessments.
Neither test outcomes nor future clinical measurements enter
candidate generation, scaling, or threshold selection.

\section{Time alignment and dose representation}
The modeling unit is a patient-history--action pair. For a decision
boundary $t$, $H_t$ contains historical observations available before
the boundary; the action input is the first interpretable documented
norepinephrine rate in $(t,t+30\text{ min}]$.
The future physiological target is summarized over
$(t+180,t+240\text{ min}]$. This separation prevents the target
window from entering the pre-decision representation.

An explicit zero rate is distinct from missing documentation.
Dose conversion requires interpretable units and, where needed,
a valid weight or drug concentration. For $\mu$g/min records, the
weight-normalized dose is the rate divided by weight in kg.
For mL/h records, a verified concentration is necessary before
normalizing the dose. Unresolvable records are excluded.

\section{Model outputs and uncertainty decomposition}
Each dynamics model predicts a mean $\mu_{mj}$ and variance
$\sigma^2_{mj}$ for physiological target $j$.
The moment-matched ensemble variance is
\begin{equation}
 v_j(H,a)=\frac{1}{M}\sum_{m=1}^{M}
 \left[\sigma^2_{mj}(H,a)+\mu_{mj}(H,a)^2\right]
 -\bar{\mu}_j(H,a)^2.
\end{equation}
The within-model variance and between-model disagreement capture
different sources of predictive spread. The gate uses between-model
disagreement, while prediction intervals use the ensemble predictive
distribution. Both quantities concern the observational model and
do not identify uncertainty in a causal treatment effect.

\section{Candidate eligibility and abstention}
For a candidate set $\mathcal{C}_t$, let
\begin{equation}
 \mathcal{C}^{+}_t=
 \{a\in\mathcal{C}_t:q(H_t,a)\geq\eta,\ u(H_t,a)\leq\tau\}.
\end{equation}
If $\mathcal{C}^{+}_t$ is empty, the output is abstention.
Otherwise, the agent refines its choice using released candidate
summaries and a validator restricts its final selection to
$\mathcal{C}^{+}_t$. Rejected candidates carry support or
disagreement reason codes instead of numerical response forecasts.
This design keeps a rejected prediction from being reused as if it
were accepted evidence.

\section{Interpretation of evaluation metrics}
Factual-transition coverage applies to the recorded action; decision
coverage applies to the existence and selection of an eligible
candidate. These denominators are distinct. Agreement is conditional
on a non-abstaining decision and therefore depends on selection.
A comparison at different coverage levels is not a paired comparison
on an identical patient subset.

Relative error reductions in the main text are calculated from the
reported point estimates. The MAP comparison is
$(9.2-7.3)/9.2=20.7\%$, and the heart-rate comparison is
$(11.6-9.6)/11.6=17.2\%$. The ablation risk reduction is
$(0.53-0.43)/0.53=18.9\%$. The difference in agreement between the
gated and ungated agents is 5.0 percentage points, rather than 5.0\%
relative improvement. These calculations do not imply statistical
significance.

\section{Prediction intervals}
The interval metric is empirical coverage of nominal 90\% prediction
intervals. The reported 87.8\% coverage for the joint gate remains
below the nominal 90\% level. Since interval widths and conditional
coverage estimates are not included in the comparison, the increase in
empirical coverage alone does not establish well-calibrated or equally
informative intervals across subsets.

\end{document}